# Simple stochastic processes behind Menzerath's Law

Jiří Milička[1]

***Abstract***

This paper revisits Menzerath's Law, also known as the Menzerath-Altmann Law, which models a relationship between the length of a linguistic construct and the average length of its constituents. Recent findings indicate that simple stochastic processes can display Menzerathian behaviour, though existing models fail to accurately reflect real-world data.

If we adopt the basic principle that a word can change its length in both syllables and phonemes, where the correlation between these variables is not perfect and these changes are of a multiplicative nature, we get bivariate log-normal distribution. The present paper shows, that from this very simple principle, we obtain the classic Altmann model of the Menzerath-Altmann Law.

If we model the joint distribution separately and independently from the marginal distributions, we can obtain an even more accurate model by using a Gaussian copula. The models are confronted with empirical data, and alternative approaches are discussed.

# 1 Introduction

Menzerath's Law, as introduced by Menzerath (1954), is a well-established relationship between the length of a construct and the average length of its constituents. For example, there is a well-documented inverse relationship between the length of words in syllables and the average length of syllables in those words. At the end of the twentieth century, Gabriel Altmann revitalized Menzerath's Law, generalizing it and describing it through equations (Altmann 1980), which is why we often refer to it now as the Menzerath-Altmann Law (MAL).

A recent study (Torre – Dębowski – Hernández-Fernández 2021) has demonstrated that even simple stochastic processes can exhibit Menzerathian behavior. However, the model presented in the study does not align with real-world data, indicating that we have yet to identify stochastic processes that can accurately model Menzerath's Law in real-world contexts. In my view, the main issue with this article (similarly to Benešová – Čech 2015) is that it focuses on processes that play a role in the production of text, while the length of constructs such as words is determined not during the production of text, but over their long evolution.

Progress was made by Milička (2023), who demonstrated that some Menzerathian datasets could be explained through regression towards the mean. Instead of focusing merely on the average lengths of constituents, he recommends examining individual constructs one by one, specifically looking at the joint distribution between the length of constructs measured in constituents and the length of constructs measured in subconstituents. For example, rather than examining how many phonemes the average syllable in two-syllable words has, it is better to look at how many words in a text have exactly two syllables and exactly five phonemes.

This may appear to be a different approach, but the Menzerath-Altmann Law can still be calculated from this joint distribution, just by summation along the y axis and then dividing the result by x. In fact, Menzerath himself measured this joint distribution in his original publication on the topic (Menzerath 1954, p. 96).

The paper by Milička (2023) on regression towards the mean is based on the idea that there is a direct proportionality between the length of a word in phonemes and its length in syllables, which is, however, slightly disturbed by noise. We can imagine that during the formation and evolution of a word, numerous forces act upon it, determining how many syllables and phonemes it has. These forces are largely independent of each other but not entirely so, resulting in an imperfect correlation between the length of a word in syllables and in phonemes. This leads to the emergence of an intercept, meaning that words with a smaller number of syllables have on average shorter syllables than the average in the entire text. Thus, if the average length of a syllable in a language is, say, 2.5 phonemes, then a two-syllable word should have 5 phonemes. However, noise disrupts this precise relationship on both axes.

This, however, is merely a crude indication of the underlying principle, as the article does not specify how exactly this noise occurs. In the given article the author promises to return to the topic and focus on the possible stochastic processes behind the joint distribution. This current paper aims to fulfil that promise.

## 2 Explanation of the classical Altmann's model by bivariate log-normal distribution

The classic Altmann model of Menzerath's Law (Altmann 1980) is represented by the equation

$$y = ax^{-b}. \quad (1)$$

This model was not derived from any specific stochastic process, and its parameters have not been given a satisfactory explanation (Kulacka 2010). However, if the joint distribution is modelled as log-normal, this explains the form of the equation very well, as we will demonstrate in this section.

It makes quite good sense to model joint distributions and marginal distributions together. For example, when a two-syllable word changes its number of phonemes from five to four during its evolution, this affects both the distribution of phonemes in words (one word with five phonemes disappears and one with four appears) and the joint distribution: the number of two-syllable words with four phonemes increases while the number of five-phoneme two-syllable words decreases. The simplest model might be the bivariate Gaussian model, whose underlying stochastic principle is quite straightforward: there are a multitude of forces that influence whether a phoneme or syllable is added to or deleted from a word, and these forces can be dependent on each other, meaning the two variables correlate (which is realistic, as the addition of a syllable in a word likely means the addition of phonemes as well).

From the bivariate Gaussian model, it is straightforward to derive the hyperbolic model of Menzerath's Law, which is described in Milička (2014). This follows simply from regression towards the mean (Galton 1886, Milička 2023):

$$y = \frac{a}{x} + b. \qquad (2)$$

By substituting $z = xy$, we then obtain a linear regression

$$z = \alpha + \beta x, \qquad (3)$$

from which an interpretation of parameters, depending on the parameters of the marginal distributions (their standard deviation $s_z$ and $s_x$ and their means $\bar{x}$ and $\bar{z}$) and the correlation between these two variables $\rho_{x,z}$, can be derived:

$$\beta = \rho_{x,z} \frac{s_z}{s_x}, \quad (4)$$

$$\alpha = \bar{z} - \beta\bar{x}. \quad (5)$$

Substituting $z = xy$ then gives us parameters for the original hyperbolic model of Menzerath's Law:

$$b = \rho_{x,xy} \frac{s_{xy}}{s_x}, \quad (6)$$

$$a = \overline{xy} - b\bar{x}. \quad (7)$$

By stating this, we do not imply that the hyperbolic model necessarily follows from a bivariate Gaussian distribution; quite the contrary, there are many bivariate distributions that result in a linear regression. While the hyperbolic model fits the data of Menzerath's Law well, the bivariate Gaussian model does not fit the empirical joint distribution very closely, and it fits the marginal distributions, such as the distribution of the number of syllables in words, even less. These are typically described by a one-displaced Poisson or log-normal distribution (Grzybek, 2007) and log-normality is present in many language-related phenomena (Torre et al. 2019).

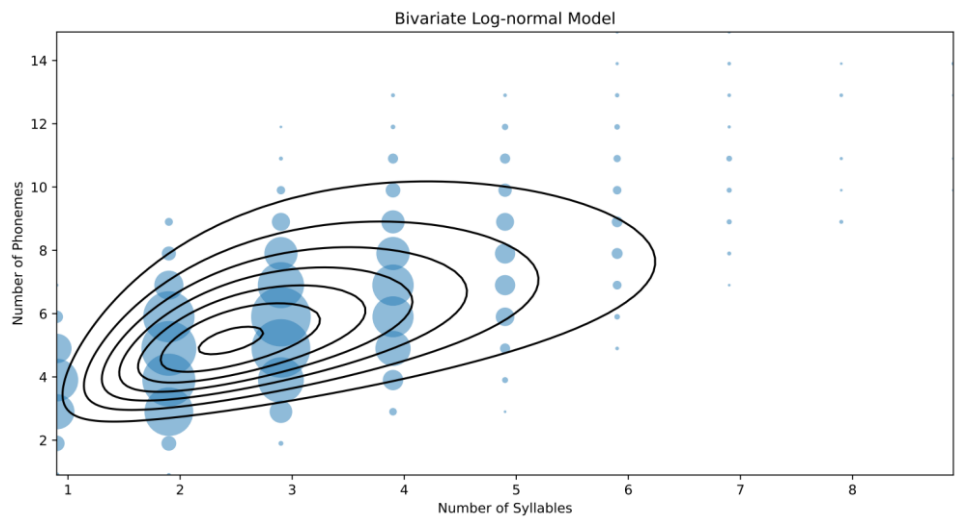

**Figure 1.** Joint distribution of number of syllables and number of phonemes modelled by log-normal bivariate distribution. Original Menzerath's data (Menzerath 1954:108).

However, we can take advantage of the fact that a bivariate log-normal distribution is relatively straightforward to achieve: simply log-transform the data along both axes and then fit a bivariate normal distribution. This joint distribution fits the empirical data at least visually quite well (see Figure 1) and, most importantly, provides an explanation for the classic Altmann equation (1). We begin again with a linear regression, this time of the log-transformed data along both axes:

$$\log z = \alpha + \beta \log x, \quad (8)$$

where the parameters can again be explained well:

$$\beta = \rho_{\log x, \log z} \frac{s_{\log z}}{s_{\log x}}, \quad (9)$$

$$\alpha = \overline{\log z} - \beta \overline{\log x}. \quad (10)$$

Then, we manipulate equation (8) to de-logarithmize it:

$$\log z = \log \alpha x^\beta, \quad (11)$$

$$z = \alpha x^\beta, \quad (12)$$

and finally, we substitute back $z = xy$, to obtain the original Menzerath-Altmann Law:

$$xy = \alpha x^\beta, \quad (13)$$

$$y = \frac{\alpha x^\beta}{x}, \quad (14)$$

$$y = \alpha x^{\beta - 1}, \quad (15)$$

$$y = ax^{-b}, \quad (16)$$

where the parameters *a* and *b* can be interpreted as:

$$b = 1 - \beta = 1 - \rho_{\log x, \log xy} \frac{s_{\log xy}}{s_{\log x}}, \qquad (17)$$

$$a = \overline{\log xy} - (1-b)\overline{\log x}. \qquad (18)$$

This distribution assumes that the underlying stochastic process is not additive but multiplicative: the probability that a seven-phoneme word will lose two phonemes is higher than the probability for a four-phoneme word to lose any. Similarly, the probability that a five-syllable word will gain another syllable is higher than for a one-syllable word. The fundamental principle still holds: we assume a large number of different forces that have a multiplicative effect on the word length.

One of the advantages of having an interpretation of the parameters is that we are not surprised when empirical data reveal a MAL with a negative parameter *b* and thus an increasing MAL curve (for instance, Motalová 2022:112,121,138 and several charts onward). This occurs when the original parameter $\beta$ is greater than 1, which can easily happen, depending on qualities of the marginal distributions and the correlation between the two variables (high correlation or low variance of *x*).

Given that the classical formula of Menzerath's Law is well-established, there is no need to empirically corroborate it in this article, for which we do not have the space.

We could take advantage of the fact that marginal distributions can be well modelled using a one-displaced Poisson distribution (Grzybek, 2007) and explore the Poisson bivariate distribution, whose properties are already described in the literature (Lakshminarayana et al. 1999, Inouye et al. 2017). However, I am not able to apply it myself and would leave it to the reader as a suggestion for further research.

## 3 *Using Gaussian Copula*

Modelling marginal distributions (e.g., the distribution of the number of syllables in words and the distribution of the number of phonemes in words) and the joint distribution simultaneously can be challenging, as it requires us to model three concepts at once. By decoupling the joint distribution from the marginal distributions, we can progress towards a more precise model. Our aim is to employ a method that allows us to be agnostic about the nature of the marginal distributions, focusing solely on the relationship between the two variables. To achieve this objective, copulas are employed.

Again, the simplest choice is the Gaussian copula, which shares a straightforward stochastic principle similar to that of the bivariate normal distribution. To fit it, knowing only the marginal distributions and the correlation coefficient between the two variables suffices. As it turns out, the Menzerath's Law model based on Gaussian Copula fits quite well, as evidenced by figures from various datasets (Figures 2 to 8). As a metric to gauge the fit, we employed the Residual Sum of Squares (RSS), because we are comparing datasets of equal length.

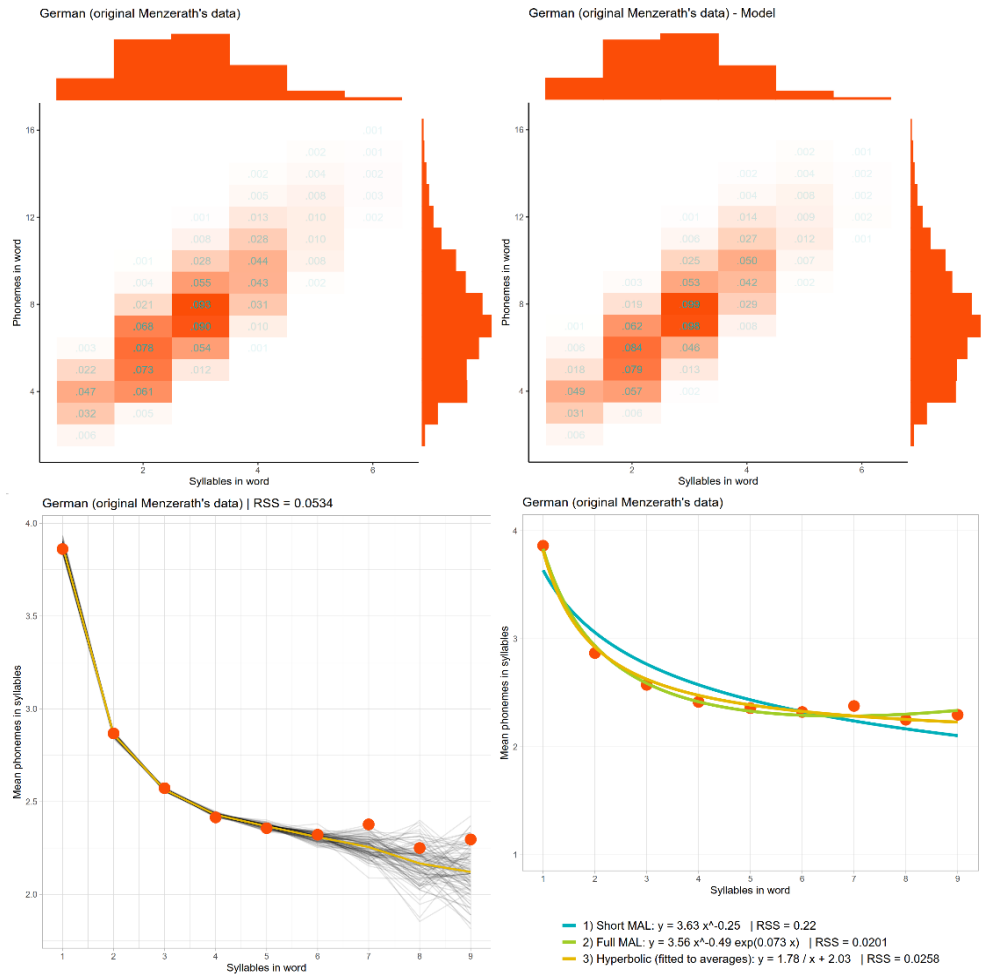

**Figure 2.** One hundred random samples from Gaussian copula to model the original Menzerath's data (Menzerath 1954:108). Top part is the whole joint distribution, bottom left is the Menzerath's law, bottom right is a comparison with the classical models.

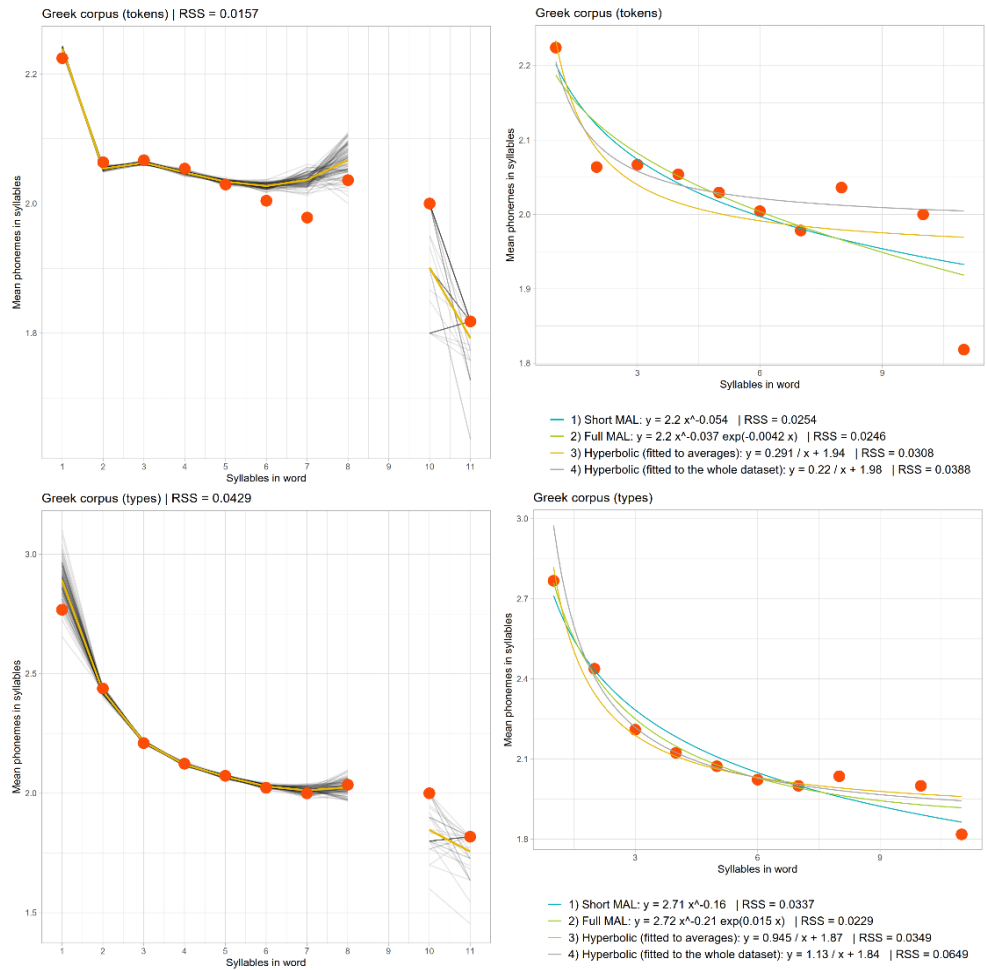

**Figure 3.** Menzerath's law on phoneme-syllable-word level in Greek (Mikros & Milička 2014).

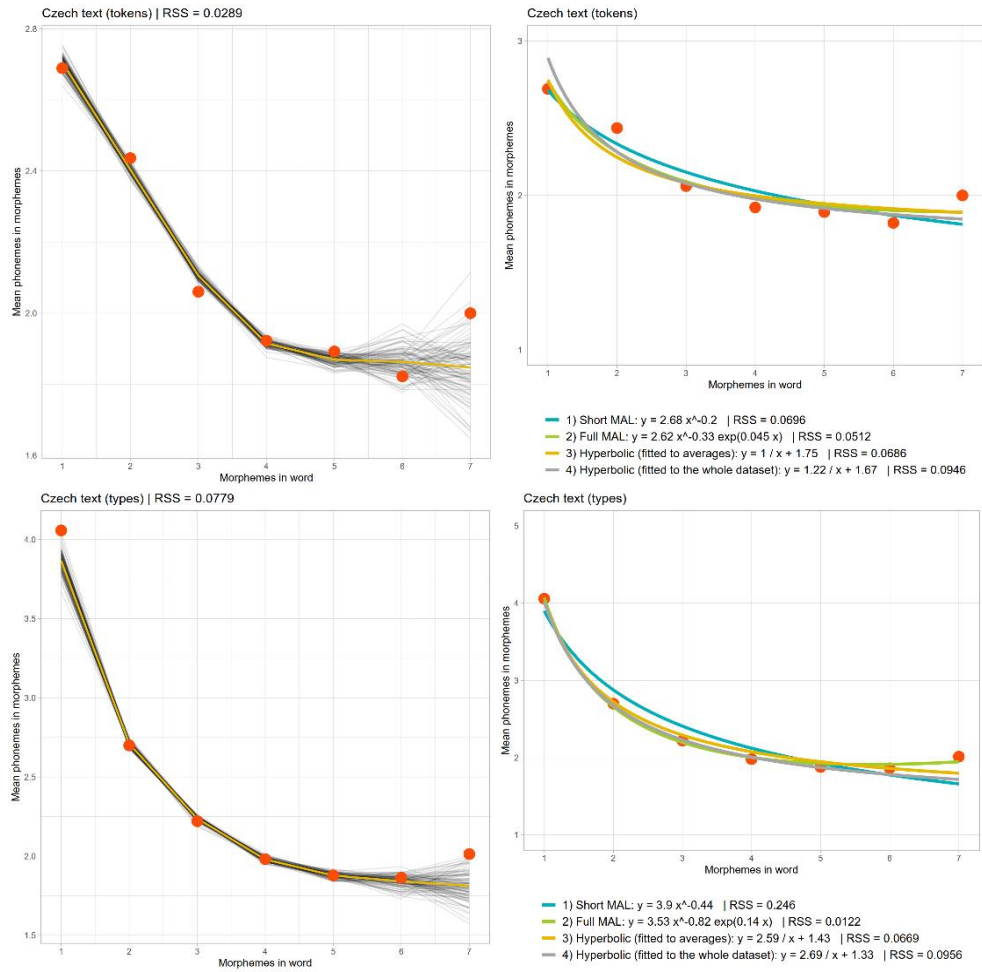

**Figure 4.** Menzerath's law on phoneme-morpheme-word level in Czech (Milička 2015).

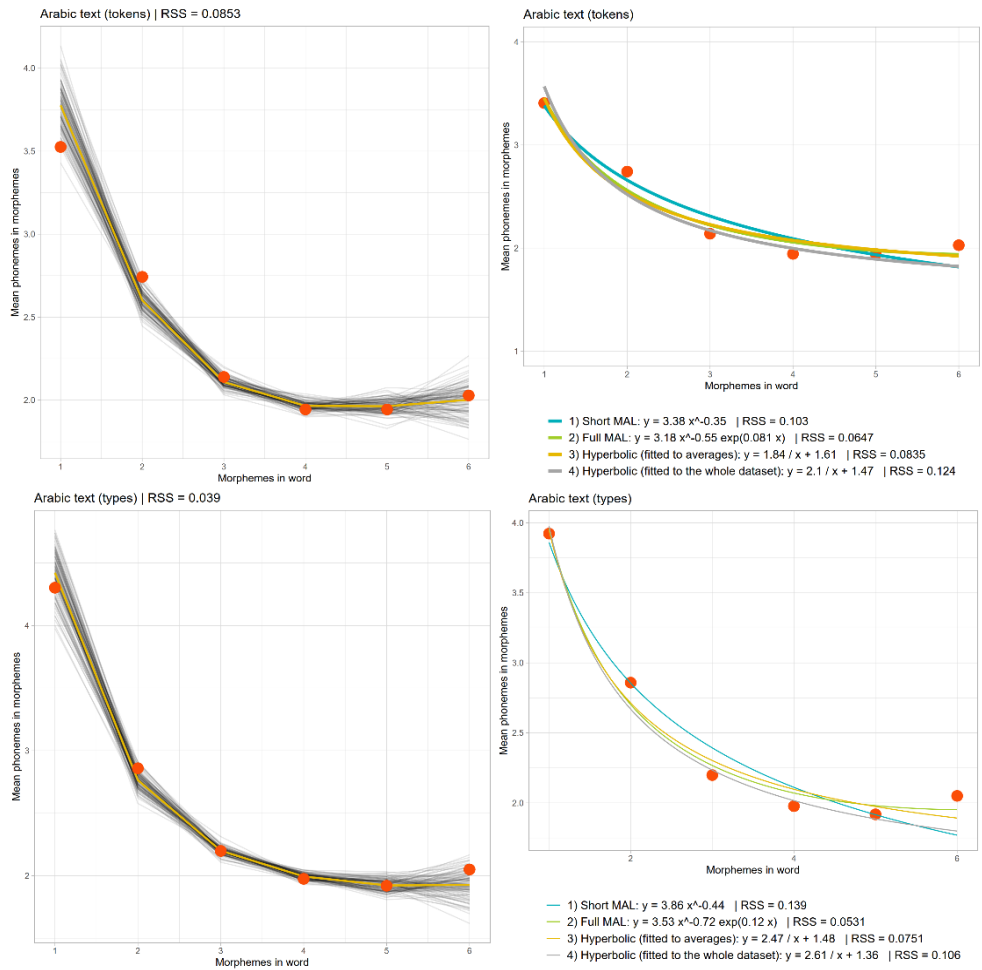

**Figure 5.** Menzerath's law on phoneme-morpheme-word level in Arabic (Milička 2015).

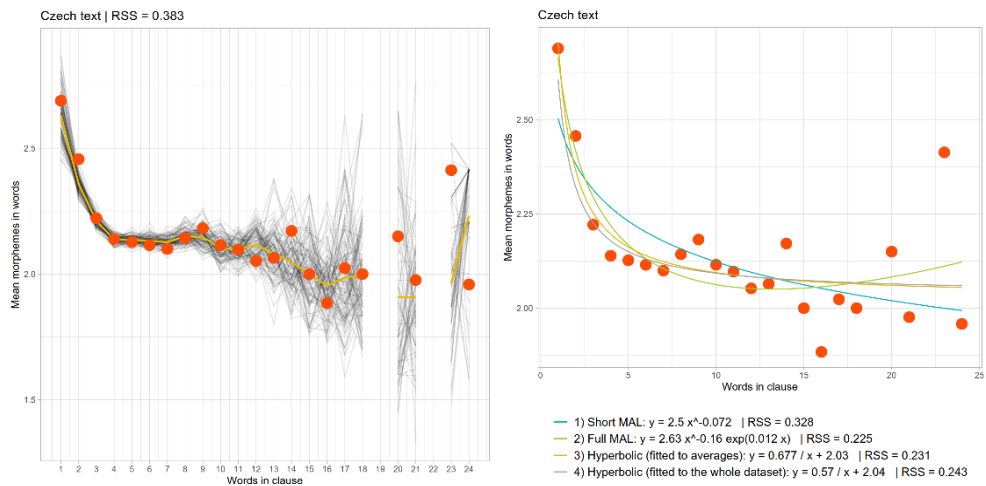

**Figure 6.** Menzerath's law on morpheme-word-clause level in Czech (Milička 2015).

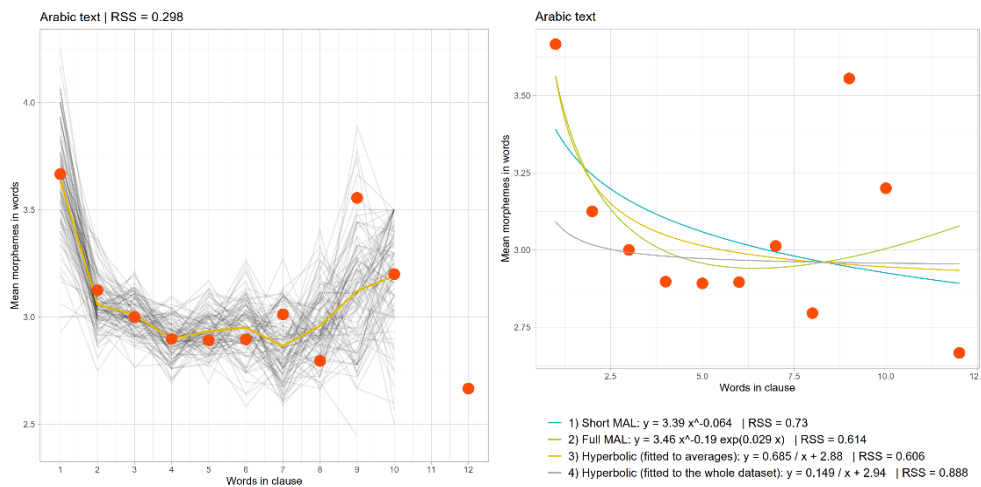

**Figure 7.** Menzerath's law on morpheme-word-clause level in Arabic (Milička 2015).

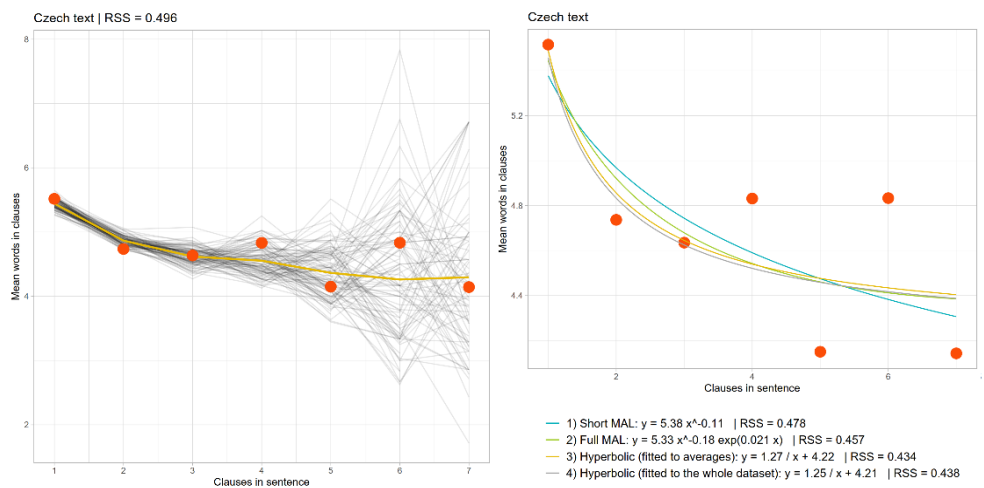

**Figure 8.** One hundred random samples from Gaussian copula to model the Menzerath's law on word-clause-sentence level in Czech (Milička 2015).

This stochastic process not only works well for Menzerath's original dataset, but also for many other datasets measured at different language levels and in various languages (e.g. Czech phoneme-morpheme-word level, see Figure 2 bottom).

As seen in Figure 7, the capability to model both increasing and decreasing trends is beneficial, as empirical data from Menzerath's Law exhibit such behaviour. Thus, formulas that can only model decreasing trends (such as the original Altmann's model or Milička's hyperbolic model) encounter difficulties in accurately reflecting the data.

The logical next step is to apply a Gaussian copula on logarithmized data: Given the good fit of the bivariate log-normal distribution discussed in the previous section, we could consider logarithmizing the entire dataset before fitting the Gaussian copula. This approach would again test the hypothesis that the stochastic process is not additive but multiplicative.

We attempted this, but it did not offer advantages; the results are usually worse.

Exploring other types of copulas, such as the Gumbel or Clayton, could also be considered. While they may perform well, finding a linguistic explanation for the stochastic process they model would be necessary. Essentially, this would bring us back to the beginning, to the year 1980: to a functioning model for which we lack an explanation.

## 4   Using segment boundaries instead of segments

As illustrated in Fig. 2, the original data collected by Paul Menzerath include empty spaces that cannot be occupied due to the definition of the relationship. For example, it is not possible to find words with three syllables and two phonemes.

This is of no concern when modelling the joint distribution by bivariate normal distributions as in section 2, since bivariate normal distributions can take negative values anyway, and there are non-zero values in regions that should be zero by definition, so this approach will never be clean. However, with copulas, it is sensible to be cautious, as the marginal distributions are fixed and cannot go into negative values, so the model actually can be done properly by considering the number of boundaries between segments (x' and y') rather than the number of segments directly.

By focusing on boundaries, we can ensure that these empty spaces from the joint distribution disappear. For instance, a word with 2 syllables and 7 phonemes has one syllable boundary and five phoneme boundaries that are not syllable boundaries.

$$x' = x - 1, \quad (19)$$

$$y' = y - x. \quad (20)$$

This transformation is simple and can be reversed at the end.

Modelling the joint distribution of boundaries of segments instead of the joint distribution of segments offers a cleaner approach (see figure 9). However, in reality, it yields worse results on all datasets where it has been tested.

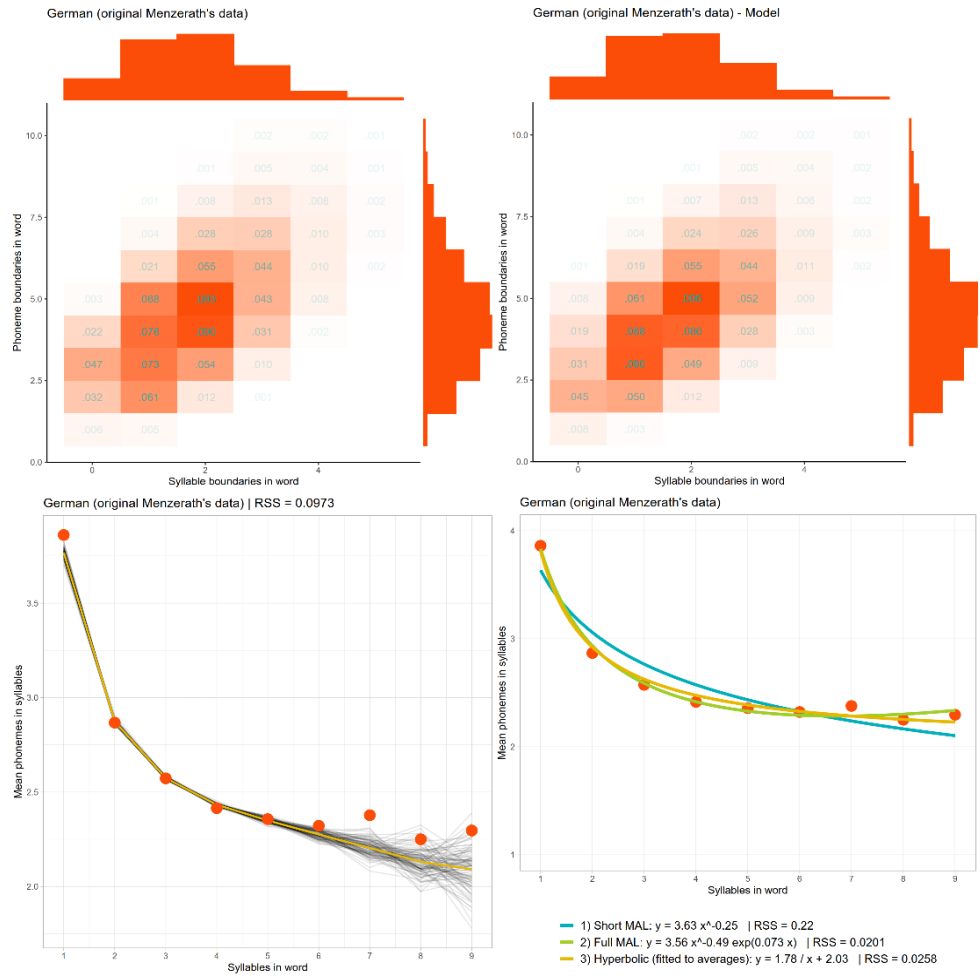

**Figure 9.** Model of the transformed Menzerath's data (Menzerath 1954:108).

## 5   *Conclusions*

This article is not exhaustive but highlights key areas for future modeling of Menzerath's Law, newly formulated as a consequence of the relationship between the length of a construct in constituents and subconstituents (for example, the number of morphemes in a given word and the number of phonemes in that word). The paper outlines two potential directions for research and poses two critical questions that need to be addressed.

The first direction resides in modelling marginal distributions and the joint distribution together to develop a unified theory of the process by which the length of constructs at different levels is determined.

The second direction lies in modelling the joint distribution separately from the marginal distributions, which is considerably easier and aligns with the reductionist approach that, despite efforts to shift science towards more holistic methods (as seen in Köhlerian synergetic linguistic, Köhler 1993), still yields fruitful results.

The first question concerns how to handle empty spaces in the joint distribution, for instance, words that cannot exist because they would have more syllables than phonemes. This article addresses this issue by modelling the number of boundaries instead of the number of segments, but as it turns out, this might not be the optimal approach. It seems that stochastic principles likely work with the original number of segments in a way that functions more effectively.

The second question is whether the stochastic process is better described as additive, multiplicative, or some amalgamation of both. Interestingly, both approaches can achieve comparably effective models, making it hard to definitively state which is better.

Take home messages from this paper are:

1) The bivariate log-normal distribution represents a linguistically plausible stochastic principle capable of modelling the length of constructs in both constituents and subconstituents. From this distribution, the classical form of Altmann's formula can be derived, and the interpretation of its parameters is relatively straightforward (equations 16 and 17).

2) When focusing solely on the joint distribution, setting aside marginal distributions, a simple Gaussian Copula proves to be an effective model for this joint distribution.

3) In any case, joint distributions provide more information than mean values. When devising a new model for Menzerath's Law, modeling the joint distribution should be a priority; the model for Menzerath's Law will naturally follow.

4) When applying Menzerath's Law parameters for practical purposes (e.g., stylometry or text profiling), using parameters from a robust model of the marginal distributions and the correlation coefficient should be considered, as it might lead to improvement.

The main reason why Menzerath's Law (MAL) is so popular within the quantitative linguistic community is due to its generality and its capacity to integrate different levels of segmentation into a unified framework, even if it has always encountered limitations at higher units (Wang & Chen 2022). However, a linguistically plausible stochastic model can vary across different levels. For instance, it can be conjectured that at the word level, the plausible stochastic process does not occur during communication but rather throughout the evolution of language: numerous factors over centuries have randomly increased or decreased word length in various units of measurement (syllables, morphemes, graphemes, or phonemes...) while the units of measurement were effected partially independently. This conjecture might not necessarily apply at the level of clauses or sentences, where the effects concurrent with the communication process probably play a more significant role. Thus, it remains to be seen whether it will be possible to identify a reasonable stochastic process that successfully covers all levels.

## 6  *Acknowledgements*

This work has been supported by Charles University Research Centre program No. 24/SSH/009.